\documentclass[conference]{IEEEtran}
\usepackage{times}

\usepackage[numbers]{natbib}
\usepackage{multicol}
\usepackage[bookmarks=true]{hyperref}
\usepackage{amsfonts}
\usepackage{amsmath}
\usepackage{xcolor}
\usepackage{graphicx}
\usepackage{lipsum}
\usepackage{booktabs}
\usepackage{cuted}
\usepackage{caption}
\usepackage{colortbl}
\usepackage{multirow}
\usepackage[most]{tcolorbox}

\newtcolorbox{greybox}{
    colback=gray!10,      
    colframe=gray!10,     
    sharp corners,        
    boxrule=0pt,          
    enhanced,             
    breakable             
}
\def\fsdf{f^\mathrm{sdf}}
\def\idt{aa}
\def\keycolor{green!50!black}
\def\valcolor{blue}

\begin{document}

\title{ShapeGen: Robotic Data Generation \\for Category-Level Manipulation}




%
\author{
\authorblockN{Yirui Wang, Xiuwei Xu, Angyuan Ma, Bingyao Yu, Jie Zhou, Jiwen Lu}
\authorblockA{Tsinghua University}
}

\def\Dsrc{D_\mathrm{src}}


\maketitle
\IEEEpeerreviewmaketitle
\begin{strip}
    \centering
    \includegraphics[width=\textwidth]{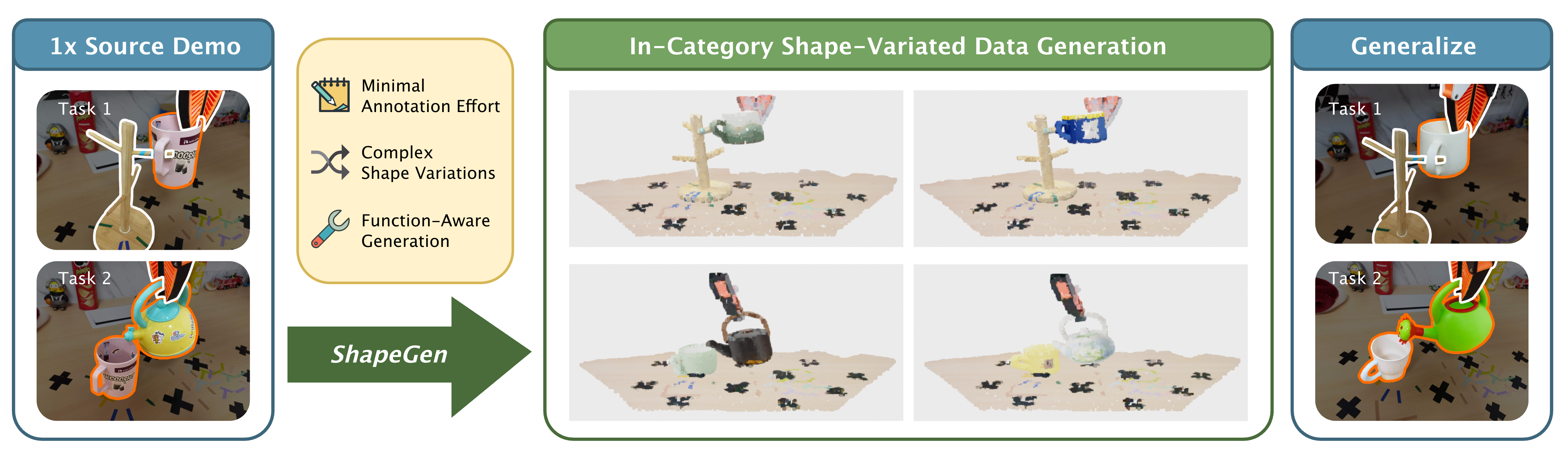}
    \captionof{figure}{\textbf{ShapeGen overview.} Given a source demo with minimal human annotation, ShapeGen automatically generates novel manipulation data with complex shape variations of manipulated objects while maintaining their functionality. Policies trained with ShapeGen-augmented data can generalize to objects of different shapes in the same category, enabling acquisition of category-level skills.}
    \label{fig:teaser}
\end{strip}

\begin{abstract}
    Manipulation policies deployed in uncontrolled real-world scenarios are faced with great in-category geometric diversity of everyday objects. In order to function robustly under such variations, policies need to work in a category-level manner, i.e. knowing how to interact with any object in a certain category, instead of only a specific one seen during training. This in-category generalizability is usually nurtured with shape-diversified training data;
    however, manually collecting such a corpus of data is infeasible due to the requirement of intense human labor and large collections of divergent objects at hand. In this paper, we propose ShapeGen, a data generation method that aims at generating shape-variated manipulation data in a simulator-free and 3D manner. ShapeGen decomposes the process into two stages: Shape Library curation and Function-Aware Generation. In the first stage, we train spatial warpings between shapes mapping points to points that correspond functionally, and aggregate 3D models along with the warpings into a plug-and-play Shape Library. In the second stage, we design a pipeline that, leveraging established Libraries, requires only minimal human annotation to generate physically plausible and functionally correct novel demonstrations. Experiments in the real world demonstrate the effectiveness of ShapeGen to boost policies' in-category shape generalizability. Project page: \url{https://wangyr22.github.io/ShapeGen/}.
\end{abstract}

\section{Introduction}

With recent advances \cite{chi2025dp, ze2024idp3, ze2024dp3, yan2025maniflow, black2024pi_0, intelligence2025pi_05, kim2024openvla}, visuomotor robotic policies are gaining more capability at being deployed in real-world manipulation scenarios, which brings to them the full complexity of everyday objects. In this context, mastering a manipulation skill requires knowing how to perform that skill to \textbf{any} object in a specific category: for example, being able to `pour water from a kettle to a mug' means being able to pour water from \textbf{any} kettle to \textbf{any} mug, regardless of their appearance: color, texture, and most importantly, shape--since the feasible action trajectories are ultimately dependent on the shapes. Such generalizability of end-to-end policies are usually cultivated by providing training data with sufficient in-category shape diversity \cite{hu2024data-scaling-law, khazatsky2024droid, o2024open, saxena2025matters, bu2025agibot}, which is usually formidable to collect in real-world for its labour-intensive nature and demand for rich collections of objects to manipulate.

Against this background, robotic data generation provides for a practical solution. Leveraging techniques such as 3D reconstruction, video inpainting and world models \cite{wang2025vggt, liu2025geometry, qian2025wristworld, zhou2024robodreamer, kerbl20233dgs, mildenhall2021nerf}, comprehensive research has been done for augmenting visual textural appearance \cite{yu2023rosie, ali2025cosmos, yuan2025roboengine, liu2025robotransfer}, spatial configuration \cite{xue2025demogen, xu2025r2rgen, zhao2025real2edit2real, yu2025real2render2real, mandlekar2023mimicgen}, viewpoint \cite{zhao2025real2edit2real} and embodiment \cite{chen2024rovi}, while leaving in-category shape mostly underexplored. Methods like RoboSplat \cite{yang2025robosplat} and ReBot \cite{fang2025rebot} approach augmenting shape of manipulated objects by replaying a task trajectory on a novel object, but are confined to simple pick-and-place-level tasks, since they do not process affordance-relevant information. To enable generation for finer-grained manipulation tasks such as mug-hanging, CP-Gen \cite{lin2025CPGen} models the skill as keypoint-trajectory constraints, i.e. keypoints on the robot or grasped object should track a reference trajectory. In the generation process, the method requires that new shapes as well as new robot actions comply with such constraints. However, CP-Gen works solely in simulation, and only supports axis-aligned scaling, which can transform explicitly keypoints to those corresponding points on the new shape, for shape variation, thus introducing far less diversity compared to complex, non-linear shape deformations in real-world scenarios.

For boosting stronger in-category shape generalizability to empower fully category-level skills for real-world deployment, we want a method that is simulator-free, works for fine-grained manipulation skills, and supports complex shape variations. We adopt a straightforward data generation idea of substituting manipulated objects with novel ones to composite data. We observe that performing fine-grained skills requires exploiting correctly the functional part of an object, and generating physically plausible novel data means its way of placement and manner of being gripped should conform to its affordances. Both requirements indicate that the generation method should be able to place the novel shape in the right place, and this can be instinctively reduced to finding the optimal alignment between the novel shape and the real object appearing in collected demo; and the alignment problem can further be reduced to finding point correspondence, for aligning transformations can be solved by minimizing a points-based cost.

Built on thes observations, in this paper, we propose a framework, ShapeGen, as a solution. We decompose the generation task into two stages: \textbf{Shape Library Curation} and \textbf{Function-Aware Generation}. In the first stage, we use geometric information to train spatial warpings that establish function-aware correspondence between 3D shapes of the same category, as the foundation for solving automatically optimal alignments.
We further aggregate a diverse collection of 3D models and trained spatial warpings into an extendable, plug-and-play Shape Library.
In the second stage, we design an effective generation pipeline that requires only minimal human annotation, provided once for each source demonstration, and leverages curated Shape Libraries to compose physically plausible and functionally correct novel manipulation data. Our system is real-to-real and time-efficient. As demonstrated in Fig. \ref{fig:teaser}, ShapeGen can generate highly realistic manipulation data. Through real-world experiments, we show the effectiveness of our system to boost shape generalizability of manipulation policies.

\section{Related Works}

\subsection{Robotic Data Generation}
Most existing researches on robotic data generation aims at generating diversified data from one or a few source demonstrations. A line of research \cite{yu2023rosie, yuan2025roboengine, liu2025robotransfer, alhaija2025cosmos1, ali2025cosmos} focuses on diversifying the textural apperance of visual observation while keeping actions and geometry of manipulated object unchanged. Another outstanding line of works \cite{mandlekar2023mimicgen, xue2025demogen, yu2025real2render2real, xu2025r2rgen, zhao2025real2edit2real} targets at diversifying spatial configuration via techniques such as Task and Motion Planning (TAMP), pointcloud editing, 3D reconstruction and 3D-controlled video generation.

There have also been works addressing shape diversity within training data. RoboSplat \cite{yang2025robosplat} uses 3D content generation model to provide a collection of data for manipulation, and adopts an off-the-shelf grasping algorithm for synthesizing grasp pose. ReBot \cite{fang2025rebot} diversify manipulated objects by replaying a human-demonstrated trajectory in simulation with different object configuration. Lacking mechanics to process object functionality and affordance, these methods can only augment data for pick-and-place-level tasks. In contrast, CP-Gen \cite{lin2025CPGen} generates shape-augmented data for fine-grained manipulation tasks in a simulator. Modelling skill segments as keypoint-trajectory constraints, given an annotated source trajectory, CP-Gen is able to adopt it to a novel shape by synthesizing an action trajectory that satisfies the same constraints as the source. However, CP-Gen can only variate the shape up to an axis-aligned scaling, since it relies on an explicit transformation to obtain corresponding keypoints on the new shapes. Thus, the level of shape complexity it introduces is not on par with complexity observed in real-world settings. 

\subsection{Inter-object Correspondence in Manipulation}

Extracting inter-object correspondence has proved useful for robotic manipulation. Usually derived from semantic feature, correspondence is widely used for transferring object affordance such as contact point for interaction \cite{ju2024roboabc, kuang2024ram}. Aside of this, \cite{wang2023d3fields, wang2023sparsedff} use similarity of feature to build planning cost for obtaining optimal trajectory. Correspondence can also be used to retarget finer-grained skills. Like the way CP-Gen \cite{lin2025CPGen} utilizes keypoint trajectory constraints, CorDex \cite{he2026cordex} transfers fingertip contact points in simulation to generate functional dexterous grasping data. MimicFunc \cite{tang2025mimicfunc} leverages an VLM for transferring functional keypoints, enabling one-shot tool usage in real-world scenarios. 

However, most methods primarily focus on discrete correspondence. Lacking global regularization methods, discrete correspondence may be inaccurate, necessitating subsequent refinement or validation \cite{he2026cordex}. In contrast with the sparse setting, DenseMatcher \cite{zhu2024densematcher} builds dense correspondence that maps vertices of one mesh to those of another, optimized under a cost that considers both semantics and regularization. Though comprehensive, the formulation of vertex correspondence still poses limitations. It is worth noticing that semantic-aware dense correspondence can be established on pure geometric foundations, as demonstrated by works \cite{park2019deepsdf, zheng2021deepimplicittemplate} targeting at category-level geometric representations. Methods built on Neural Descriptor Fields (NDFs) \cite{simeonov2022ndf, cheng2023nodtamp} also exploits this observation by training category-level descriptors with pure occupancy supervision. Based on the same assumption and taking a step further to dense correspondence, ShapeGen uses geometric clues to prepare and aggregate dense inter-object correspondence for an efficient data generation system.

\begin{figure*}[htbp]
    \centering
    \includegraphics[width=\linewidth]{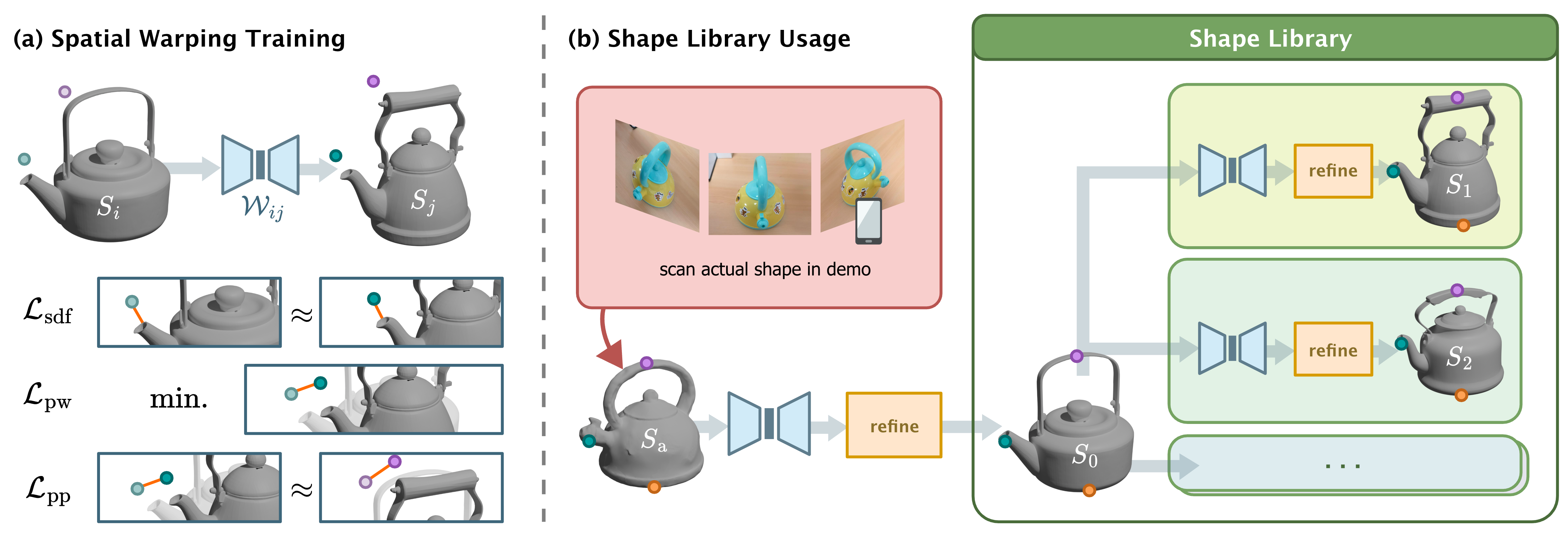}
    \caption{\textbf{Shape Library curation overview}. (a) Spatial warpings are trained with geometric supervision including SDF loss, point-wise regularization loss and point-pair regularization loss. (b) Shape Library is constructed by aggregating warpings stemming from a common template. Given a scanned shape, the Library can be used in a plug-and-play manner by solely training an additional warping. When calculating the composite warping, extra refining steps are added to mitigate error.}
    \label{fig:shape-library}
\end{figure*}

\section{Method}

\subsection{Problem Statement}

A visualmotor policy is a mapping $\pi: \mathcal{O} \to \mathcal{A}$ from observation space to action space. To train a policy fulfilling a specific task, usually a corpus of data $\mathcal{D} = \{D_1, ..., D_N\}$ is prepared, with each $D = (d_0, d_1, ..., d_{L-1})$ a sequence of observation-action pairs $d_t = (o_t, a_t)$. In order for a policy to obtain certain \textbf{generalizability}, the data should hold the corresponding type of \textbf{diversity}. Our proposed method, ShapeGen, aims at augmenting the shape diversity of task-relevant objects via data generation, thus reducing the human effort as well as requirement for possessing shape-diversified assets for collecting data. Given a source demonstration $\Dsrc$ with minimal human annotation, ShapeGen automatically substitutes task-relevant objects in $\Dsrc$ with \textit{objects of the same category but different shapes} (\textit{shapes} for short) in a physically plausible and functionally correct manner, yielding a large corpus of novel training data.

\textbf{Overview.}
In this work, we capture raw observation with a calibrated RGB-D camera, and train policies taking uncolored pointcloud observation. To generate data, ShapeGen first curate a Shape Library for each relevant category of object that contains not only 3D shapes, but also spatial warpings to them from a common template shape, detailed in \ref{subsec:shape-library}. Given a sequence of raw observation-action pairs, we ask an annotator to recognize keypoints on the actual objects used, designate methods for inter-shape alignment, then uses the established Shape Library to automatically calculate optimal alignment and correct action, detailed in \ref{subsec:generation}.

\subsection{Shape Library Curation}\label{subsec:shape-library}

\begin{figure*}[htbp]
    \centering
    \includegraphics[width=\linewidth]{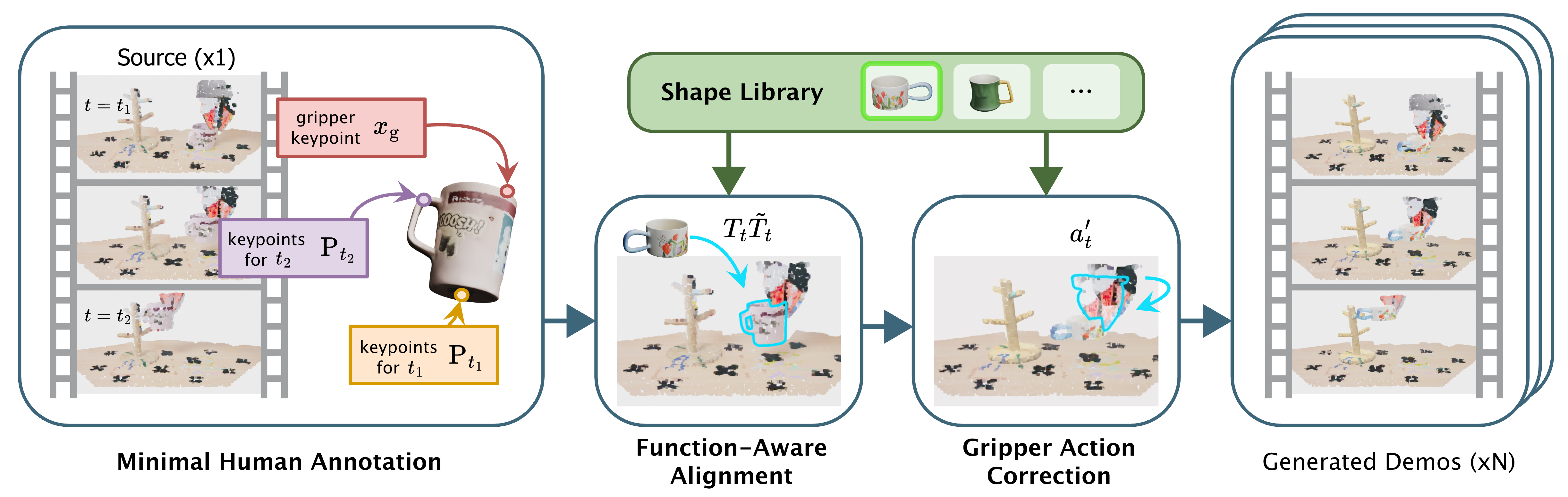}
    \caption{\textbf{Data generation pipeline.} A human annotator is only required to annotate once for each demo. With 3D shapes and warping networks provided by Shape Libraries, function-aware alignment and gripper action correction can be performed in a fully automated manner.}
    \label{fig:data-generation}
\end{figure*}


In this section, we aim to develop an automatic mechanism to, given the actual object appearing in source demo and keypoints attached to it, find on every shapes in a Library the points corresponding to them. This will enable point-based alignment in the generation pipeline. Such correspondence should be function-aware, e.g. given a mug and a point on inner side of its handle, the mapped point should too be on the inner side of mug handles. In \ref{subsubsec:spatial-warping-training}, we introduce method for finding such correspondence between two shapes. In \ref{subsubsec:shape-library-construction}, we introduce the way to aggregate correspondence into a plug-and-play Shape Library.

\subsubsection{Spatial Warping Training}\label{subsubsec:spatial-warping-training}

Given two 3D shapes $S_i$ and $S_j$ belonging to the same category, we seek to find dense function-aware correspondence, i.e. a spatial warping $\mathcal{W}_{ij}: \mathbb{R}^3 \to \mathbb{R}^3$ between them. We notice that, across divergent in-category shapes, functional correspondence is generally derived from geometric correspondence. This is natural, since function of an object as well as of its specific parts themselves are usually deduced from the very geometric shape of the object.
Based on this observation, we propose to establish $\mathcal{W}_{ij}$'s via geometric clues. Since the final correspondence may be complex, we use neural networks to fit $\mathcal{W}$'s, and use pure geometric information for supervision.

Specifically, we adopt Signed Distance Function (SDF) as the supervising signal. Given a 3D shape $S$ and a point $\boldsymbol{x}\in\mathbb{R}^3$, the absolute value of $\mathrm{SDF}(\boldsymbol{x}, S)$ represents the distance from $\boldsymbol{x}$ to the surface of $S$, while the sign indicates whether $\boldsymbol{x}$ is inside or outside of $S$. To establish correspondence, we hope that given $\boldsymbol{x}_i$ near surface of $S_i$, $\mathcal{W}_{ij}$ sends it to $\mathcal{W}_{ij}(\boldsymbol{x}_i)$ such that
\begin{equation}
    \mathrm{SDF}(\boldsymbol{x}_i, S_i) = \mathrm{SDF}(\mathcal{W}_{ij}(\boldsymbol{x}_i), S_j)
\end{equation}

Note that in this way, $\mathrm{SDF}(\boldsymbol{x}_i, S_i) =0$ indicates $\mathrm{SDF}(\mathcal{W}_{ij}(\boldsymbol{x}_i), S_j)=0$, which means points on surface are mapped to points on surface. Additional requirements from non-surface points may provide more comprehensive geometric supervision. In order to train $\mathcal{W}_{ij}$ in an end-to-end manner, we first fit nerual SDF model for $S_j$ to obtain spatial-differentiable SDF function, noted $f^\mathrm{sdf}_j$. Then, we fix $f^\mathrm{sdf}_j$'s and train $\mathcal{W}_{ij}$. The primary training loss is
\begin{equation}
    \mathcal{L}_\mathrm{sdf} = \Vert f^\mathrm{sdf}_i(\boldsymbol{x}_i, S_i) - f^\mathrm{sdf}_j(\mathcal{W}_{ij}(\boldsymbol{x}_i), S_j)\Vert_1
\end{equation}
In which, letting $y:=\mathcal{W}_{ij}(\boldsymbol{x}_i)$ and $z:= f^\mathrm{sdf}_j(y)$ and denote by $\theta$ parameters of $\mathcal{W}_{ij}$, the gradiant can be calculated as
\begin{equation}
    \frac{\partial\mathcal{L}_\mathrm{sdf}}{\partial\theta} = \frac{\partial\mathcal{L}_\mathrm{sdf}}{\partial z} \cdot \frac{\partial z}{\partial y} \cdot \frac{\partial y}{\partial \boldsymbol{x}_i}
\end{equation}
Where the spatial derivative of SDF function $\partial z/{\partial y} = \partial f^\mathrm{sdf}_j/\partial y$ can be easily obtained, as $f^\mathrm{sdf}_j$ is in form of a neural network.
Following \cite{zheng2021deepimplicittemplate}, we also add point-wise and point-pair regularization losses to avoid getting overly deformed warpings:
\begin{equation}
    \mathcal{L}_\mathrm{pw} = h(\Vert\mathcal{W}_{ij}(\boldsymbol{x}_i) - w_i \Vert_2)
\end{equation}
\begin{equation}
    \mathcal{L}_\mathrm{pp} = \max\left( \frac{\Vert \Delta\boldsymbol{x}_i^{(1)} - \Delta\boldsymbol{x}_i^{(2)} \Vert_2}{\Vert \boldsymbol{x}_i^{(1)} - \boldsymbol{x}_i^{(2)} \Vert_2} - \varepsilon, 0 \right)
\end{equation}
Where $h(\cdot)$ is a Huber kernel, superscripts in $\boldsymbol{x}_i^{(1)}, \boldsymbol{x}_i^{(2)}$ denote different samples in a batch, and $\Delta\boldsymbol{x}_i := \mathcal{W}_{ij}(\boldsymbol{x}_i) - \boldsymbol{x}_i$. The final training loss is a weighted sum:
\begin{equation}
    \mathcal{L} = \mathcal{L}_\mathrm{sdf} + \alpha_\mathrm{pw} \mathcal{L}_\mathrm{pw} + \alpha_\mathrm{pp}\mathcal{L}_\mathrm{pp}
\end{equation}
as illustrated in Fig. \ref{fig:shape-library} (a).

\subsubsection{Shape Library Construction and Usage}\label{subsubsec:shape-library-construction}

We assemble divergent shapes and warping functions between them into a plug-and-play Shape Library, as illustrated in Fig. \ref{fig:shape-library} (b). For a fixed category, we pick out a template shape $S_0$, and train warpings $\mathcal{W}_{0j}$ from it to all other shapes. This allows for adding new shape to the Library in $\mathcal{O}(1)$ time, and also enables `plugging' the Library to any shape $S_\mathrm{a}$: by training warping $\mathcal{W}_{\mathrm{a}0}$ from $S_\mathrm{a}$ to $S_0$ and compositing the warpings we seamlessly get $\mathcal{W}_{\mathrm{a}j} = \mathcal{W}_{0j} \circ \mathcal{W}_{\mathrm{a}0}$. In the data generation process to be introduced, $S_\mathrm{a}$ is a scanned model of actual object being manipulated, and in this way we get information for substituting it with any shape from the Library. When calculating the composite warping function, we add an refining optimization step to find the nearest point on surface to hone outputs of $\mathcal{W}_{0j}$ and $\mathcal{W}_{\mathrm{a}0}$, for mitigating error caused by imperfect $\mathcal{W}$'s.
\begin{equation}
    \mathrm{refine}(\boldsymbol{x}, S) = \mathop{\arg\min}_{\boldsymbol{y} \in S}\Vert \boldsymbol{x} - \boldsymbol{y}\Vert_2
    \label{eq:refining-step}
\end{equation}
\begin{equation}
    \mathcal{W}_{\mathrm{a}j}(\boldsymbol{x}) = \mathrm{refine}(\mathcal{W}_{0j}(\mathrm{refine}(\mathcal{W}_{\mathrm{a}0}(\boldsymbol{x}), S_0)), S_j)
\end{equation}
The optimization problem in \ref{eq:refining-step} can be easily solve approximately via SDF network and spatial gradients, as detailed in appendix \ref{subsec-apdx:warping-refinement}.

\subsection{Function-Aware Generation}\label{subsec:generation}

In this section, built upon curated Shape Libraries, we introduce the data generation pipeline, in which task-relevant objects in $\Dsrc$ are substituted with shapes from the Libraries. 
For simplicity, we assume here that there is only 1 relevant object, with shape $S_\mathrm{a}$, in the source demonstration, and we wish to substitute it with $S_j$ from the library, in a realistic manner.
Before running the pipeline, we scan a 3D model $S_\mathrm{a}$ of the actual shape used using a mobile application. Then, beginning with raw RGB-D observation (with known camera intrinsics) and robot arm action, we first run an out-of-the-box, model-based object tracking method \cite{wen2024foundationpose} to obtain the object pose trajectory $\boldsymbol{T}_\mathrm{a} = (T_0, ..., T_{L-1}), T_t \in \mathrm{SE}(3)$ of $S_\mathrm{a}$. Then, we ask a human annotator to provide a small amount of annotation, detailed in \ref{subsubsec:min-human-annotation}. After this, for each shape in the Library, we perform first function-aware alignment (\ref{subsubsec:function-aware-alignment}) and then gripper action correction (\ref{subsubsec:gripper-action-correction}), getting the modified action sequence. At last, using the calculated alignments and poses, we generate novel observation (\ref{subsubsec:novel-obs-generation}), yielding a new demonstration episode on a different object shape. The generation pipeline is outlined in Fig. \ref{fig:data-generation}.

\subsubsection{Minimal Human Annotation}\label{subsubsec:min-human-annotation}

Given the source demonstration, we ask a human annotator to provide annotations $\{(t_i, \mathbf{P}_{t_i}, G_{t_i})\}$, where $\mathbf{P}_{t_i} = (\boldsymbol{x}_n \in \mathbb{R}^3)$ is a orderd set of points fixed to task-relevant objects appearing in the demonstration, $t_i$ a timestamp, and $G_{t_i}$ a alignment cost function. The purpose of such annotations is to guide the subsequent aligning step for putting a new shape in the `right' place. Take the task \verb|hang_mug| for example: in the hanging process, the upper corner of the mug handle should be on the exact spot in order for inserting it onto the rack, thus the annotator can take $t_i=$ the timestamp when hanging begin, and $\mathbf{P}_{t_i}$ the singleton set of a point on the upper corner of the mug handle. $G_{t_i}$ can be a simple square error, indicating that the keypoint position should be preserved.  Usage of $G_{t_i}$ and the specific aligning process will be introduced in the subsequent section. Aside of this, we also ask the annotator to recognize a gripper keypoint $\boldsymbol{x}_\mathrm{g}$ that represent the position of object gripped by the gripper. The aforementioned annotation need only be performed \textbf{once} for each source demo, taking only about 1 minute. Annotaion process is detailed in appendix \ref{subsec-apdx:annotation-procedure}.

\subsubsection{Function-Aware Alignment}\label{subsubsec:function-aware-alignment}

With annotations provided, we solve for an optimal alignment transformation $\tilde T_{t_i}$ via
\begin{equation}
    \tilde T_{t_i} = \mathop{\arg\min}_{T \in \mathrm{SE}(3)} G_{t_i}\left(\mathbf{P}_{t_i}, T\left[\mathcal{W}_{\mathrm{a}j}[\mathbf{P}_{t_i}]\right]\right)
\end{equation}
Where $\mathcal{W}_{\mathrm{a}j}[\mathbf{P}_{t_i}]$ is the image of ordered set $\mathbf{P}_{t_i}$ under function $\mathcal{W}_{\mathrm{a}j}$, and $T[\mathcal{W}_{\mathrm{a}j}[\mathbf{P}_{t_i}]]$ its subsequent image under transformation $T$. As a result, $\tilde T_{t_i}$ is the transformation that optimally aligns points on $S_j$ to their corresponding points on $S_\mathrm{a}$, and henceforth aligns $S_j$ to $S_\mathrm{a}$. In our experiments, $G_{t_i}$ are usually square distances, and $\mathbf{P}_{t_i}$ singletons. In these cases, where the optimal solution is not unique, we impose an additional constraint that the rotation component of $\tilde T_{t_i}$ be identity.\footnote{The shapes $S_\mathrm{a}$ and $S_j$ are aligned in advance to canonical orientation as a preprocessing step, like in most 3D shape datasets.}

Upon obtaining $\tilde T_{t_i}$'s, we interpolate them to get the alignment sequence $(\tilde T_{0}, ..., \tilde T_{L-1})$, and furthur calculate the desired object pose trajectory of $S_j$ as
\begin{equation}
    \boldsymbol{T}_j = (T_0 \tilde T_0, ..., T_{L-1} \tilde T_{L-1})
\end{equation}
In some cases, we not only want $S_\mathrm{a}$ and $S_j$ aligned in the shared object space, but also need the attached points $\mathbf{P}_{t_i}$ and $\mathcal{W}_{\mathrm{a}j}[\mathbf{P}_{t_i}]$ aligned in the same manner to an anchor point attached to another variable shape: e.g. when performing the task \verb|pour_water| we want the kettle mouths aligned to a point on the mug rim, which itself may be transformed elsewhere. Under these circumstances, we calculate similarly a sequence of alignments $\hat T_{t}$ in camera space, and get
\begin{equation}
    \boldsymbol{T}_j = (\hat T_0T_0 \tilde T_0, ..., \hat T_{L-1}T_{L-1} \tilde T_{L-1})
\end{equation}

\subsubsection{Gripper Action Correction}\label{subsubsec:gripper-action-correction}

As the size as well as exact position of gripped part varies between shapes, in order for the novel demonstration to be physically correct, the action of gripper should also be modified accordingly. Using $\boldsymbol{T}_\mathrm{a}$ and $\boldsymbol{T}_j$ we get trajectories of the annotated gripper keypoint $\boldsymbol{x}_\mathrm{g}$ as well as its corresponding point $\mathcal{W}_{\mathrm{a}j}(\boldsymbol{x}_\mathrm{g})$ in camera space, from which we calculate a sequence of translation-only transformation $\boldsymbol{T}_\mathrm{g} = (T^\mathrm{g}_0, ..., T^\mathrm{g}_{L-1})$ that best aligns the points and henceforth transforms the observed gripper to a new position where it grips the substituted shape well. We represent the gripper action $a_t$ as a $\mathrm{SE}(3)$ transformation from end-effector frame to robot base frame, and calibrate beforehand a transformation from camera frame to robot base frame $T_\mathrm{c2b}$; consequently, we get modified action
\begin{equation}
    a_t' = T_\mathrm{c2b} T^\mathrm{g}_t T_\mathrm{c2b}^{-1} a_t
\end{equation}
We require that $T^\mathrm{g}_t$ be translation-only mainly for two reasons. First, this avoid the need to render the arm and gripper from a new angle, because the visual fidelity is usually well-preserved under a small translation. Second, in this way we only need the rotation component of $T_\mathrm{c2b}$ for calculating $a_t'$ (detailed in appendix \ref{subsec-apdx:action-correction}), hence reducing the negative impact of imprecise hand-eye calibration.

\subsubsection{Novel Observation Generation}\label{subsubsec:novel-obs-generation}

After calculating $\boldsymbol{T}_j$ and $\boldsymbol{T}_\mathrm{g}$, we composite the new observation by manipulating representation in 3D space. Using a video segmentation model \cite{ravi2024sam2}, we obtain masks for each task-relevant object. Also, assuming a static environment, we capture an RGB-D image of the empty workspace. We obtain mask of the robotic arm by first segmenting the foreground regions via comparing depth values and then extracting masks of all other objects. We use $\boldsymbol{T}_\mathrm{g}$ to directly transform the segmented arm pointcloud, and use $\boldsymbol{T}_j$ to put the 3D model of $S_j$ in correct place. In our experiments, we adopt a 3D policy that takes pointcloud as observation, so we directly sample points from composited scene with a camera-aware 3D post-processing method proposed in \cite{xu2025r2rgen}. However, our system also works for 2D policies if the scene is rendered into images.

\section{Experiments}

In this part, we conduct experiments to validate the effectiveness of ShapeGen method. In \ref{subsec:experiment-generalizability}, we verify that ShapeGen can boost the in-category shape generalizability of manipulation policies across 4 manipulation tasks. In  \ref{subsec:experiment-realism}, we validate that data generated by ShapeGen hold sufficient realism. In \ref{subsec:efficiency-analysis}, we analyze the efficiency of ShapeGen over manual data collection. In \ref{subsec:comparison-with-fmatching}, we compare our spatial warping method with naive feature matching. In \ref{subsec:visualization}, we provide visualizations for correspondence established via Shape Libraries and data generated by our system.

\begin{table*}[htbp]
  \centering
  \setlength{\tabcolsep}{3pt}
  \caption{\textbf{Results of ShapeGen for shape generalizability.} \#i denote the i-th novel object instance, tested once for each of 5 spatial configurations. Success rate is reported.}
\resizebox{0.98\linewidth}{!}{
    \begin{tabular}{l|ccccc|ccccc|ccccc|ccccc}
    \toprule
          & \multicolumn{5}{c|}{\textbf{hang\_mug}} & \multicolumn{5}{c|}{\textbf{hang\_mug\_hard}} & \multicolumn{5}{c|}{\textbf{serve\_kettle}} & \multicolumn{5}{c}{\textbf{pour\_water}} \\
\cmidrule{2-21}          & \#1   & \#2   & \#3   & \#4   & \cellcolor[rgb]{ .804,  .953,  .976}total & \#1   & \#2   & \#3   & \#4   & \cellcolor[rgb]{ .804,  .953,  .976}total & \#1   & \#2   & \#3   & \#4   & \cellcolor[rgb]{ .804,  .953,  .976}total & \#1   & \#2   & \#3   & \#4   & \cellcolor[rgb]{ .804,  .953,  .976}total \\
    \midrule
    \textbf{source} & 0/5   & 0/5   & 0/5   & 1/5   & 1/20  & 1/5   & 0/5   & 0/5   & 0/5   & 1/20  & 1/5   & \textbf{5/5}   & 1/5   & 0/5   & 7/20  & \textbf{5/5}   & \textbf{2/5}   & 2/5   & 2/5   & 11/20 \\
    \textbf{+ShapeGen} & \textbf{3/5}   & \textbf{3/5}   & \textbf{1/5}   & \textbf{2/5}   & \textbf{9/20}  & \textbf{3/5}   & \textbf{4/5}   & \textbf{2/5}   & \textbf{1/5}   & \textbf{10/20} & \textbf{4/5}   & 3/5   & \textbf{5/5}   & \textbf{3/5}   & \textbf{15/20} & \textbf{5/5}   & 1/5   & \textbf{3/5}   & \textbf{3/5}   & \textbf{12/20} \\
    \bottomrule
    \end{tabular}%
}
  \label{tab:main-experiment}%
\end{table*}%

\begin{figure*}[htbp]
    \centering
    \includegraphics[width=\linewidth]{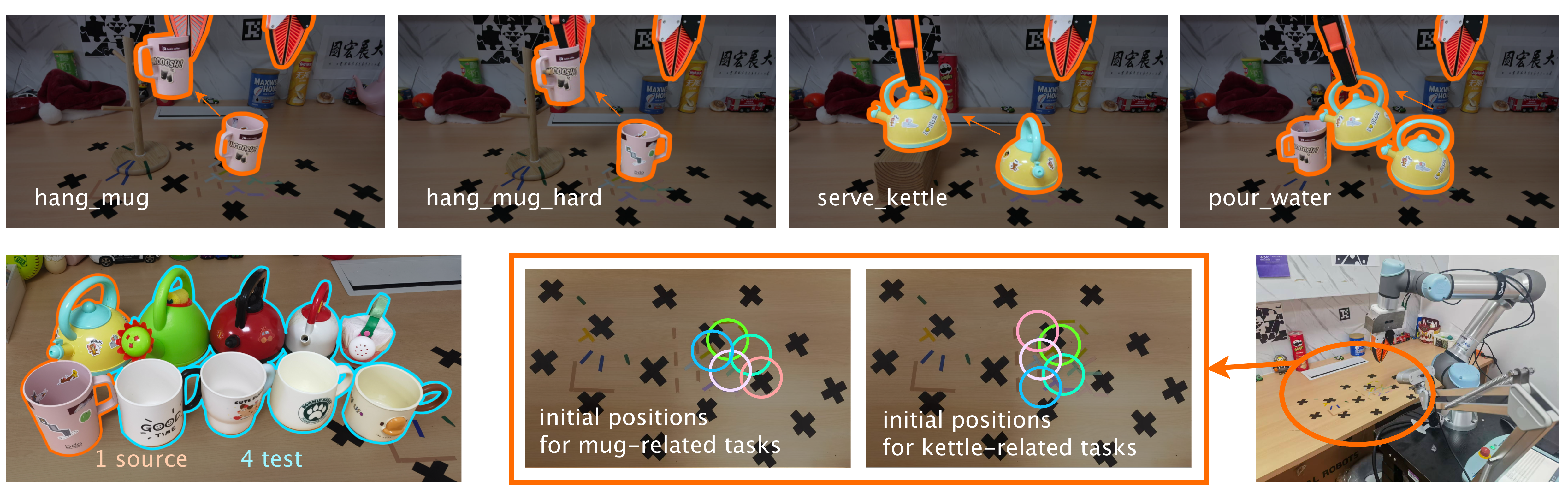}
    \caption{\textbf{Visualization of tasks.} We conduct real-world experiments on 4 tasks requiring exploitation of object functionality. For each task, 5 demos are collected with the same object instance and different spatial configurations. Tests are conducted on novel object instances unseen in either source or generated data.}
    \label{fig:tasks}
\end{figure*}

\subsection{Implementation Details}

\subsubsection{ShapeGen and Manipulation Policy}
We utilize the efficient library Instant-NGP \cite{muller2022instantNGP} to fit neural SDF networks. We adopt the LSTM architecture used in DIT \cite{zheng2021deepimplicittemplate} with an additional residual connection for our spatial warping networks. To extend Shape Libraries, we collect Internet-source images and use SAM3D \cite{chen2025sam3d} to generate 3D models. We scan 3D models of manipulated objects with the mobile application RealityComposer, segment video observation with SAM2 \cite{ravi2024sam2} and track objects with FoundationPose \cite{wen2024foundationpose}. To validate our method, we adopt the policy architecture ManiFlow \cite{yan2025maniflow} with uncolored pointcloud input for eliminating the influence of colors and focusing majorly on shape. Data generation and all network trainings are performed on a single RTX A6000 GPU. Details on spatial warping training and policy training are provided in appendix \ref{subsec-apdx:warping-training} and \ref{subsec-apdx:warping-refinement}.

\subsubsection{Hardware Setup}
We collect data and deploy policies on a single-arm fixed-base robot platform. The platdorm consists of a 7-DoF UR5 robotic arm with a parallel jaw gripper. An ORBBEC femto bot RGB-D camera is fixed to the base.

\subsection{Shape Generalizability of Policy}\label{subsec:experiment-generalizability}

To verify that ShapeGen-augmented data can boost in-category shape generalizability of manipulation policies, we selected 4 tasks as illustrated in Fig. \ref{fig:tasks} concerning 2 categories of everyday objects, \verb|mug| and \verb|kettle|. For every task, 5 demonstrations are collected via teleoperation with same object instances and different spatial configurations. Leveraging ShapeGen, we generate 15 novel demos from each source demo. We train ManiFlow \cite{yan2025maniflow} policies from source demos only (\textbf{source}) and from both source and generated data (\textbf{+ShapeGen}). To evaluate their in-category generalizability, we conduct tests on object instances unseen in either source or generated data, placed at approximately the same positions as objects in source demos to decouple shape generalizability with spatial generalizability. For each unseen object instance, one test is carried out for each spatial configuration, since we empirically find that results are highly repeatable given same initial setup (detailed in appendix \ref{subsec-apdx:SR-variance}).

\begin{figure}[t]
    \centering
    \includegraphics[width=\linewidth]{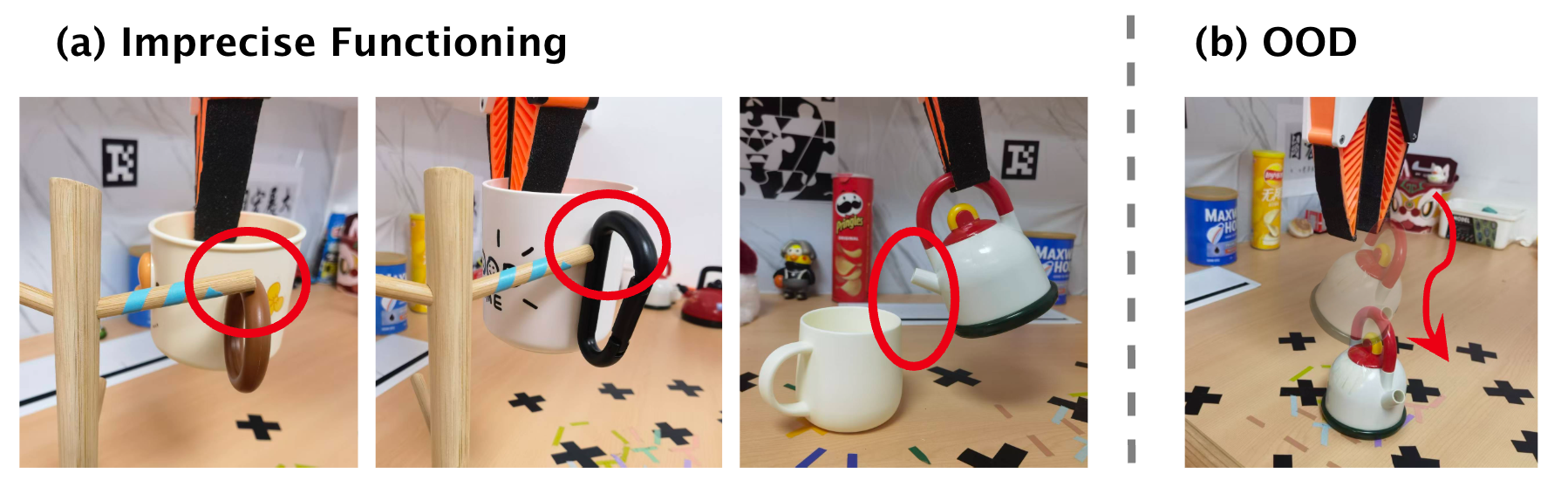}
    \caption{\textbf{Typical failure cases.} Policies commonly suffer from imprecise exploitation of objects' functions and out-of-distribution effect. For example, the mug being manipulated cannot be inserted onto the rack without collision, and the kettle being gripped is frequently dropped amidst execution.}
    \label{fig:fail-cases}
\end{figure}

\begin{table*}[htbp]
  \centering
  \setlength{\tabcolsep}{2.5pt}
  \caption{\textbf{System runtime analysis.} Typical time cost for each system components is given.}
  \resizebox{0.98\linewidth}{!}{
    \begin{tabular}{cc|cc|cc|c}
    \toprule
    \rowcolor[rgb]{ .345,  .882,  .125} \multicolumn{2}{c|}{\textbf{shape library curation}} & \multicolumn{2}{c|}{\cellcolor[rgb]{ .78,  1,  .576}\textbf{data collection preparation}} & \multicolumn{2}{c|}{\cellcolor[rgb]{ .804,  .953,  .976}\textbf{data collection}} & \cellcolor[rgb]{ .075,  .875,  .992}\textbf{data generation} \\
    train nerual SDF $T_\mathrm{sdf}$ & train warping $T_\mathrm{wrp}$ & scan $S_a$ $T_\mathrm{scan}$ & train W network $T_\mathrm{wrp}$ & collect demo $T_\mathrm{demo}$ & annotate demo $T_\mathrm{ann}$ & generate $T_\mathrm{gen}$ \\
    \midrule
    5s    & 3min  & 5min  & 3min  & 1min  & 1min  & 20s \\
    \bottomrule
    \end{tabular}%
    }
  \label{tab:runtime-analysis}%
\end{table*}%

\begin{figure*}
    \centering
    \includegraphics[width=\linewidth]{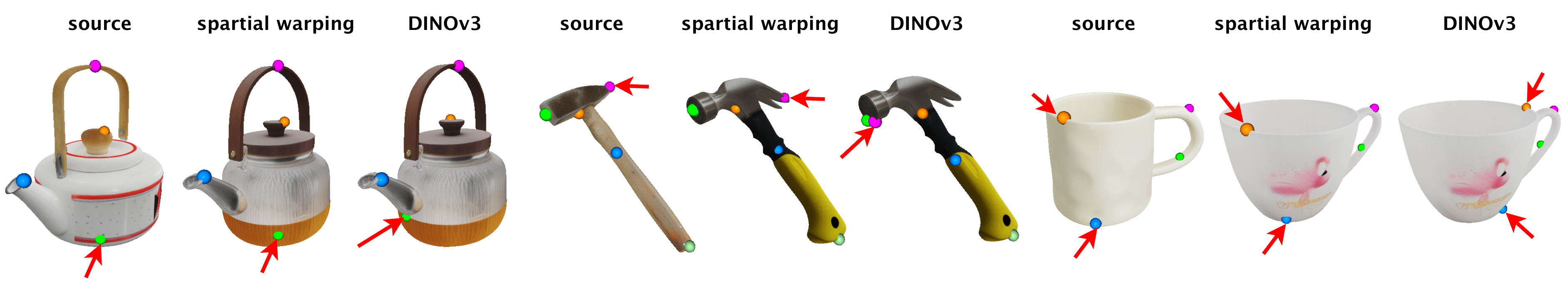}
    \caption{\textbf{Qualitative comparison with feature-matching method.} Feature-matching is prone to imprecise single-point matches; examples are highlighted with red arrows. Corresponding points are marked with the same color.}
    \label{fig:compare-dino}
\end{figure*}

As shown in Table \ref{tab:main-experiment}, ShapeGen significantly enhances generalizability of policies for the mug-hanging tasks and also brings performance gain for kettle-related tasks, the former being considerably more difficult since placing the mug onto rack requires more precise alignment. Fig. \ref{fig:fail-cases} summarizes typical failure cases. Policies trained solely on source demos most commonly suffer from imprecise functioning, in which cases failure is caused by an imprecise exploitation of objects' function: for instance, in mug-hanging tasks, they fail to insert mug handles into the rack without collision; or, in the task \verb|serve_kettle|, the spout is not aligned with the mug's rim. These are also the problems leading to failure when simply replaying the action trajectory on a shape-variated object, because due to shape variation, the same trajectory cannot precisely align the functional parts. Another common reason for failure is out-of-distribution effect. For example, in the task \verb|serve_kettle|, policy trained on only source data frequently drops the kettle amidst execution. By enriching the training corpora with functionally-correct shape-augmented data, ShapeGen enables policies to better exploit functions of unseen objects. Also, the diversity injected into training data helps policies reduce O.O.D. failures.

\subsection{Realism of Generated Data}\label{subsec:experiment-realism}

\begin{table}[t]
  \centering
  \caption{\textbf{Results of data realism.} ShapeGen generates feasible trajectory and enables policies to learn successfully from them.}
  \resizebox{\linewidth}{!}{
    \begin{tabular}{c|cc|cc}
    \toprule
    & \multicolumn{2}{c|}{\textbf{task/replay}} & \multicolumn{2}{c}{\textbf{task/policy}} \\
    \cmidrule{2-5}          & \textbf{hang\_mug} & \textbf{hang\_mug\_hard} & \textbf{hang\_mug} & \textbf{hang\_mug\_hard} \\
    \midrule
    \textbf{source} & 0/5   & 0/5   & 0/5   & 0/5 \\
    \textbf{ShapeGen} & 5/5   & 5/5   & 5/5   & 5/5 \\
    \bottomrule
    \end{tabular}%
    }
  \label{tab:shapegen-fidelity}%
\end{table}%

We verify the realism of generated data by running ShapeGen pipeline with scanned model of a test object \verb|mug_white| used as the model from a Shape Library, generating demos containing only the object scanned. Then, we train policies with these generated data (\textbf{policy-ShapeGen}) and compare them with policies trained on source demos only (\textbf{policy-source}), tested on the scanned object. We also directly replay the collected as well as generated action trajectories (\textbf{replay-source}, \textbf{replay-ShapeGen}) on the scanned object, yielding results in Table \ref{tab:shapegen-fidelity}. While both the source trajectory and the source policy fail altogether, trajectories generated by ShapeGen can fulfill correctly the task and enable policies to learn from them. This shows that: on the one hand, ShapeGen produces functionally correct trajectories; on the other hand, observation generated by ShapeGen are sufficiently realistic so that the real-world observations during execution are in-distribution for policies trained.

\subsection{Efficiency Analysis}\label{subsec:efficiency-analysis}

\begin{figure*}
    \centering
    \includegraphics[width=\linewidth]{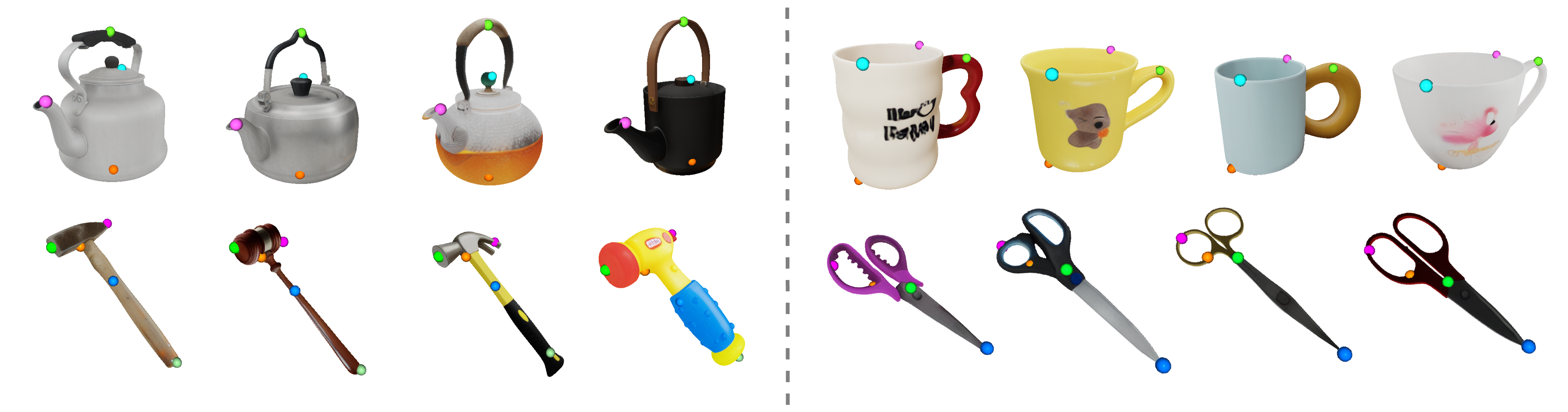}
    \caption{\textbf{Visualization of Shape Libraries.} Visualizations of categories \texttt{kettle}, \texttt{mug}, \texttt{hammer} and \texttt{scissors} are provided. In each category, points of the same color are mapped from a same point on the common template shape.}
    \label{fig:vis-library}
\end{figure*}

\begin{figure*}
    \centering
    \includegraphics[width=\linewidth]{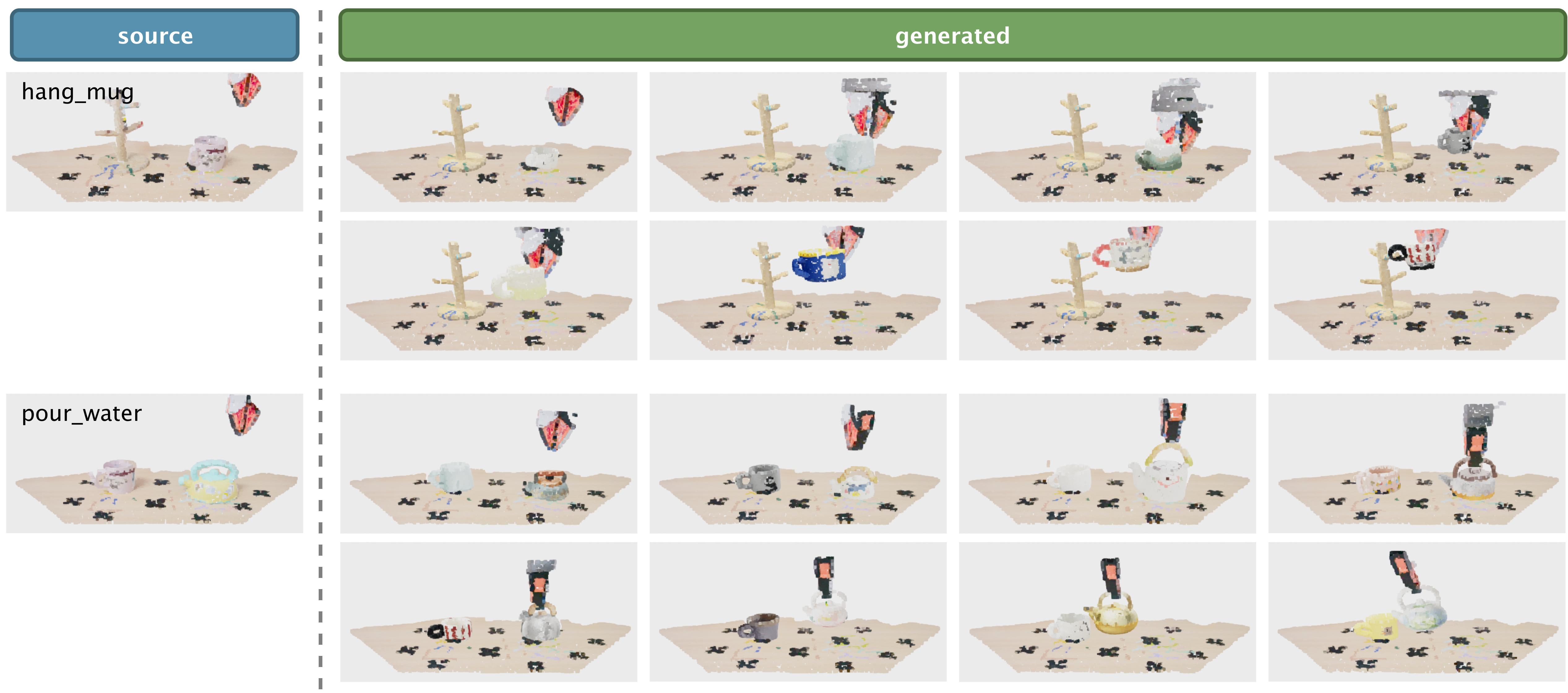}
    \caption{\textbf{Visualization of generated data.} In the generated observations, novel objects are placed correctly on the table in the start and manipulated correctly during execution to fulfill their functionalities; also, the gripper is translated accordingly to the correct location for grasping and manipulation.}
    \label{fig:vis-generated}
\end{figure*}

ShapeGen provides significant efficiency gain for increasing in-category zero-shot success rate for tasks concerning a fixed category of objects. Table \ref{tab:runtime-analysis} gives typical time requirements for running each component of the ShapeGen system. Specifically, ShapeGen generates data at about 15 fps, hence $T_\mathrm{gen} = 20\,\mathrm{s}$ being a conservative estimation since source episodes contain usually less than 300 frames for our single-skill tasks. Assuming only one category is concerned for simplicity, if we wish to generate data for $N_\mathrm{shape}$ shapes for each of $N_\mathrm{src}$ source demos (yielding a total of $N_\mathrm{shape} \cdot N_\mathrm{src}$ novel demonstrations), the overall time cost will be
\begin{equation}
\begin{split}
    T(N_\mathrm{shape}, N_\mathrm{demo}) =& (T_\mathrm{sdf} + T_\mathrm{wrp}) \cdot N_\mathrm{shape} + \\
    & (T_\mathrm{scan} + T_\mathrm{wrp}) \cdot 1 + \\
    & (T_\mathrm{demo} + T_\mathrm{ann}) \cdot N_\mathrm{demo} +\\
    & T_\mathrm{gen} \cdot N_\mathrm{shape} \cdot N_\mathrm{demo}
\end{split}
\end{equation}
With $T_\mathrm{gen}$ being the dominant coefficient when $N_\mathrm{shape} \cdot N_\mathrm{demo}$ grows. In contrast, if we collect all data via teleoperation, the time cost will be
\begin{equation}
    T'(N_\mathrm{shape}, N_\mathrm{demo}) = T_\mathrm{demo} \cdot N_\mathrm{shape} \cdot N_\mathrm{demo} + t(N_\mathrm{shape})
\end{equation}
Where $t(N_\mathrm{shape})$ is the time cost for collecting $N_\mathrm{shape}$ different objects. Notably $T_\mathrm{gen} < T_\mathrm{demo}$, making data generation by ShapeGen considerably more time-efficient than manual data collection.

\subsection{Comparison with Feature Matching}\label{subsec:comparison-with-fmatching}

We compare qualitatively single-point correspondences extracted by our spatial warping network and a naive feature-matching method. For the latter, we use the state-of-the-art image feature extraction model DINOv3 \cite{simeoni2025dinov3} and aggregate a featured pointcloud from rendered images, as detailed in appendix \ref{subsec-apdx:feature-matching}. As shown in Fig. \ref{fig:compare-dino}, lacking global regularization when extracting correspondence, feature-matching is prone to imprecise matches, hence not robust enough for integration into a fully autonomous data generation system. In contrast, our spatial warping method considers both local and global geometry, producing high-quality single-point correspondence especially for points on rather `plain' regions such as mug rims and lower parts of kettles.

\subsection{Visualizations}\label{subsec:visualization}

In this section, we provide more visualizations. Fig. \ref{fig:vis-library} shows point correspondence extracted by trained spatial warping networks. Trained solely under geometric supervision, spatial warpings in Shape Libraries can satisfactorily establish function-aware point correspondence for different shapes. This enables our pipeline to map automatically annotated keypoints on the actual shape $S_\mathrm{a}$ to each shape in the Library, automating the alignment and generation pipeline. Fig. \ref{fig:vis-generated} gives rendered pointcloud images of generated observations. In the generated data, novel object shapes are placed correctly on the table in the start, and manipulated correctly by the gripper to fulfill their functionalities.

\section{Conclusion} 
\label{sec:conclusion}

In this paper, we present ShapeGen, a robotic data generation framework designed to enhance the in-category shape generalizability of manipulation policies. Starting from a single source demonstration, ShapeGen can seamlessly substitute objects with diverse in-category shapes and modify the corresponding action trajectories correctly. This allows for automated synthesis of large-scale diversified manipulation data while requiring only a one-minute human annotation on the initial source demo. At its core, we develop an geometry-based, function-aware curation method to construct plug-and-play Shape Libraries with neural-based dense correspondences between object pairs. Notably, this library is highly scalable, supporting integration of new shapes in $\mathcal{O}(1)$ time. Leveraging established correspondences, in the data generation pipeline, ShapeGen composites novel 3D point clouds observation by substituting the actually manipulated object with various shapes under proper alignment, and transforms gripper trajectories accordingly. Extensive experiments across multiple representative tasks demonstrate that ShapeGen produces high-quality, diverse demonstrations, significantly bolstering the out-of-distribution generalization of state-of-the-art manipulation policies.

\textbf{Potential Limitations.} Although ShapeGen injects complex and flexible geometric diversity into training data, the full potential of this framework has not been exhaustively demonstrated within the current experimental scope. Future work will focus more on intricate tasks that require for policies a deeper understanding of functional parts of objects being manipulated, and further validate the potential of ShapeGen to scale up training data more drastically. Our framework does not support deformation objects, because the complex dynamics normally requires either training a dynamics model or using a physics simulator and is out of the scope of our current research; however, ShapeGen can be extended to support articulated objects via a set of engineering efforts:  (a) adopt a mesh reconstruction method that supports articulation; (b) train part-level Shape Libraries and then aggregate the parts under articulation constraints. We plan to implement this in subsequent work as an extension. 

\bibliographystyle{plainnat}
\bibliography{references}

\clearpage
\appendices

\section{Implementation Details}

\subsection{Warping Training}\label{subsec-apdx:warping-training}

We adopt the LSTM network architecture used in DIT \cite{zheng2021deepimplicittemplate} with an additional residual connection as the spatial warping architecture. Training inputs are sampled on a $256\times256\times 256$ grid on the cube $[-0.05, 1.05]^3$ given that the 3D shape is normalized to be within the unit cube. We discard points with groundtruth SDF values greater than 0.2 to ensure that the warping focuses mainly on points near the surface. We train all warpings using Adam and DIT-like learning rate scheduling, with batch size $8196$ for a total of 2 epochs. Training each warping takes about 5 minutes on a single RTX A6000 GPU.

\subsection{Policy Training and Deployment}\label{subsec-apdx:policy-training-and-deployment}

We use ManiFlow \cite{yan2025maniflow} that takes robot state with $N=8192$ uncolored points observation as input. We discard points that are over $D_{\max} = 0.8$ metres from the origin of camera frame before sampling pointcloud from depth image. We take action prediction horizon $T_p = 16$, observation horizon $T_o = 2$, and execute all actions in each action chunk at 6 Hz before inferring new actions in deployment. All policies are trained on a single RTX A6000 GPU using AdamW (learning rate $1 \times 10^{-4}$, betas $(0.9, 0.95)$, weight decay $1 \times 10^{-3}$) with batch size $256$ for 16k iterations regardless of dataset size. Training each policy takes about 5 hours.

\subsection{Annotation Procedure}\label{subsec-apdx:annotation-procedure}

We ask an annotator to provide an \verb|annotation.json| for each source demonstration. Here is an example:

\begin{greybox}
\small
\texttt{
\{ \\
\textcolor{gray!10}{\idt}\textcolor{\keycolor}{"objects"}: \{ \\
\textcolor{gray!10}{\idt\idt}\textcolor{\keycolor}{"mug\_pink"}: \{ \\
\textcolor{gray!10}{\idt\idt\idt}\textcolor{\keycolor}{"category"}: \textcolor{\valcolor}{"mug"}, \\
\textcolor{gray!10}{\idt\idt\idt}\textcolor{\keycolor}{"gripped"}: \textcolor{\valcolor}{true}, \\
\textcolor{gray!10}{\idt\idt\idt}\textcolor{\keycolor}{"gripper\_keypoint"}: [ \\
\textcolor{gray!10}{\idt\idt\idt\idt}[ \\
\textcolor{gray!10}{\idt\idt\idt\idt\idt}\textcolor{\valcolor}{-0.03592510148882866},\\
\textcolor{gray!10}{\idt\idt\idt\idt\idt}\textcolor{\valcolor}{0.10562200099229813},\\
\textcolor{gray!10}{\idt\idt\idt\idt\idt}\textcolor{\valcolor}{0.023572899401187897}\\
\textcolor{gray!10}{\idt\idt\idt\idt}]\\
\textcolor{gray!10}{\idt\idt\idt}],\\
\textcolor{gray!10}{\idt\idt\idt}\textcolor{\keycolor}{"functionals"}: [\\
\textcolor{gray!10}{\idt\idt\idt\idt}\{\\
\textcolor{gray!10}{\idt\idt\idt\idt\idt}\textcolor{\keycolor}{"tstamp"}: \textcolor{\valcolor}{0},\\
\textcolor{gray!10}{\idt\idt\idt\idt\idt}\textcolor{\keycolor}{"keypoints"}: \{\\
\textcolor{gray!10}{\idt\idt\idt\idt\idt\idt}\textcolor{\keycolor}{"mode"}: \textcolor{\valcolor}{"simple"},\\
\textcolor{gray!10}{\idt\idt\idt\idt\idt\idt}\textcolor{\keycolor}{"points"}: [\\
\textcolor{gray!10}{\idt\idt\idt\idt\idt\idt\idt}[\\
\textcolor{gray!10}{\idt\idt\idt\idt\idt\idt\idt\idt}\textcolor{\valcolor}{-0.001282079960219562},\\
\textcolor{gray!10}{\idt\idt\idt\idt\idt\idt\idt\idt}\textcolor{\valcolor}{0.0028379999566823244},\\
\textcolor{gray!10}{\idt\idt\idt\idt\idt\idt\idt\idt}\textcolor{\valcolor}{0.003494200063869357}\\
\textcolor{gray!10}{\idt\idt\idt\idt\idt\idt\idt}]\\
\textcolor{gray!10}{\idt\idt\idt\idt\idt\idt}]\\
\textcolor{gray!10}{\idt\idt\idt\idt\idt}\}\\
\textcolor{gray!10}{\idt\idt\idt\idt}\},\\
\textcolor{gray!10}{\idt\idt\idt\idt}\{\\
\textcolor{gray!10}{\idt\idt\idt\idt\idt}\textcolor{\keycolor}{"tstamp"}: \textcolor{\valcolor}{100},\\
\textcolor{gray!10}{\idt\idt\idt\idt\idt}\textcolor{\keycolor}{"keypoints"}: \textcolor{\valcolor}{"ditto"}\\
\textcolor{gray!10}{\idt\idt\idt\idt}\},\\
\textcolor{gray!10}{\idt\idt\idt\idt}\{\\
\textcolor{gray!10}{\idt\idt\idt\idt\idt}\textcolor{\keycolor}{"tstamp"}: \textcolor{\valcolor}{115},\\
\textcolor{gray!10}{\idt\idt\idt\idt\idt}\textcolor{\keycolor}{"keypoints"}: \{\\
\textcolor{gray!10}{\idt\idt\idt\idt\idt\idt}\textcolor{\keycolor}{"mode"}: \textcolor{\valcolor}{"simple"},\\
\textcolor{gray!10}{\idt\idt\idt\idt\idt\idt}\textcolor{\keycolor}{"points"}: [\\
\textcolor{gray!10}{\idt\idt\idt\idt\idt\idt\idt}[\\
\textcolor{gray!10}{\idt\idt\idt\idt\idt\idt\idt\idt}\textcolor{\valcolor}{-0.0020681601017713547},\\
\textcolor{gray!10}{\idt\idt\idt\idt\idt\idt\idt\idt}\textcolor{\valcolor}{0.08404900133609772},\\
\textcolor{gray!10}{\idt\idt\idt\idt\idt\idt\idt\idt}\textcolor{\valcolor}{-0.049066800624132156}\\
\textcolor{gray!10}{\idt\idt\idt\idt\idt\idt\idt}]\\
\textcolor{gray!10}{\idt\idt\idt\idt\idt\idt}]\\
\textcolor{gray!10}{\idt\idt\idt\idt\idt}\}\\
\textcolor{gray!10}{\idt\idt\idt\idt}\}\\
\textcolor{gray!10}{\idt\idt\idt}]\\
\textcolor{gray!10}{\idt\idt}\}\\
\textcolor{gray!10}{\idt}\},\\
\textcolor{gray!10}{\idt}\textcolor{\keycolor}{"other\_foreground\_objects"}: [\textcolor{\valcolor}{"tree"}]\\
\}\\
}
\end{greybox}
Generally, the annotator only needs to provide \verb|"gripper_keypoint"| and several entries in \verb|"functionals"|. \verb|"gripper_keypoint"| should be a point in the gripped part on $S_\mathrm{a}$ . As for the latter, in each entry, \verb|"tstamp"| stands for timestamp $t_i$, \verb|"points"| the points $\mathbf{P}_{t_i}$, and \verb|"mode"| a key designating one of several alignment cost functions $G_{t_i}$'s we have implemented. Here, \verb|"simple"| stands for the simple square error (plus a restriction that both shapes are aligned to the canonical orientation), where the optimal alignment is the translation-only transformation that aligns the center of $\mathcal{W}_{\mathrm{a}j}[\mathbf{P}_{t_i}]$ to that of $\mathbf{P}_{t_i}$. \verb|"ditto"| means using the same keypoints entry as the previous one. The annotation provided here requires to align the mug by a point on its bottom strictly from $t_1=0$ to $t_1=100$ (interpolation is actually performed between two identical alignments), interpolate between two alignments from $t_2=100$ to $t_3=115$, and align by a point on its handle strictly from $t_3=115$ on.

The point coordinate can be obtained via clicking on the 3D model in Blender. We use a simple script to get coordinate of the clicked point.

\subsection{Warping Refinement}\label{subsec-apdx:warping-refinement}

\begin{figure}
    \centering
    \includegraphics[width=0.8\linewidth]{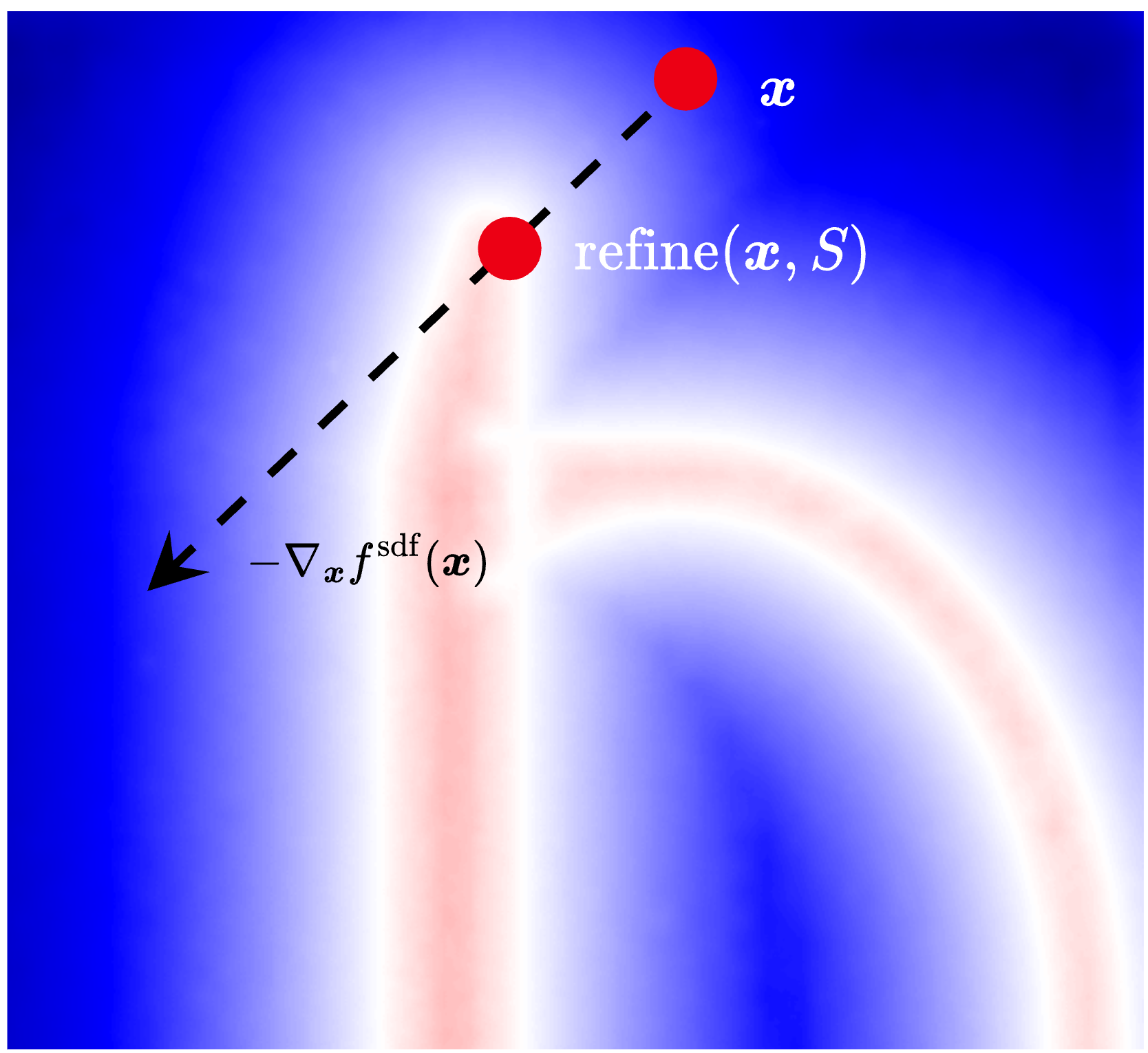}
    \caption{\textbf{Refining step.} A Line Search is done along the negative derivative direction to find a point on surface.}
    \label{fig:refine}
\end{figure}

We use a refining step,
\begin{equation}
    \mathrm{refine}(\boldsymbol{x}, S) = \mathop{\arg\min}_{\boldsymbol{y} \in S}\Vert \boldsymbol{x} - \boldsymbol{y}\Vert_2
\end{equation}
to mitigate error caused by imperfect $\mathcal{W}$ when calculating composite warping functions $\mathcal{W}_{\mathrm{a} j} = \mathcal{W}_{0 j} \circ \mathcal{W}_{\mathrm{a}0}$. To find the nearest point $\boldsymbol{y}$ on surface of $S$ efficiently, we use a simple Line Search method over the gradient direction:
\begin{equation}
\begin{split}
    t^\ast = \min& \Big\{ t\in \mathbb{R}^+ : \\
    &\fsdf\big(\boldsymbol{x} - t \cdot \nabla_{\boldsymbol{x}}\fsdf(\boldsymbol{x}) \cdot \mathrm{sgn}( \fsdf(\boldsymbol{x} ))\big) = 0\Big\}
\end{split}
\end{equation}
\begin{equation}
    \mathrm{refine}(\boldsymbol{x}, S) = \boldsymbol{x} - t^\ast \cdot \nabla_{\boldsymbol{x}} \fsdf(\boldsymbol{x}) \cdot \mathrm{sgn}( \fsdf(\boldsymbol{x} ))
\end{equation}
which is illustrated in Fig. \ref{fig:refine}.

\subsection{Action Correction}\label{subsec-apdx:action-correction}

In this section, we detail the calculation of
\begin{equation}
    a_t' = T_\mathrm{c2b} T^\mathrm{g}_t T_\mathrm{c2b}^{-1} a_t
\end{equation}
Where $T_\mathrm{c2b}$ is the $\mathrm{SE}(3)$ transformation matrix from robot base frame to camera frame, $a_t, a_t'$ end-effector pose in robot base frame, and $T_t^\mathrm{g}$ a translation-only transformation in camera frame. Note that letting
\begin{equation}
    T_\mathrm{c2b} = 
    \begin{bmatrix}
    \boldsymbol{R}_\mathrm{c2b} & \boldsymbol{t}_\mathrm{c2b} \\
    \boldsymbol{0} & 1
\end{bmatrix}, \,
T_t^\mathrm{g} = 
\begin{bmatrix}
    \boldsymbol{I} & \boldsymbol{t} \\
    \boldsymbol{0} & 1
\end{bmatrix}
\end{equation}
we get
\begin{equation}
\begin{split}
    T_\mathrm{c2b} T^\mathrm{g}_t T_\mathrm{c2b}^{-1} &=
    \begin{bmatrix}
    \boldsymbol{R}_\mathrm{c2b} & \boldsymbol{t}_\mathrm{c2b} \\
    \boldsymbol{0} & 1
    \end{bmatrix}
    \begin{bmatrix}
    \boldsymbol{I} & \boldsymbol{t} \\
    \boldsymbol{0} & 1
    \end{bmatrix}
    \begin{bmatrix}
    \boldsymbol{R}_\mathrm{c2b}^\top & -\boldsymbol{R}_\mathrm{c2b}^\top\boldsymbol{t}_\mathrm{c2b} \\
    \boldsymbol{0} & 1
    \end{bmatrix}\\
    &=
    \begin{bmatrix}
    \boldsymbol{R}_\mathrm{c2b} & \boldsymbol{t}_\mathrm{c2b} + \boldsymbol{R}_\mathrm{c2b}\boldsymbol{t}\\
    \boldsymbol{0} & 1
    \end{bmatrix}
    \begin{bmatrix}
    \boldsymbol{R}_\mathrm{c2b}^\top & -\boldsymbol{R}_\mathrm{c2b}^\top\boldsymbol{t}_\mathrm{c2b} \\
    \boldsymbol{0} & 1
    \end{bmatrix}\\
    &=
    \begin{bmatrix}
    \boldsymbol{I} & -\boldsymbol{t}_\mathrm{c2b} +\boldsymbol{t}_\mathrm{c2b} + \boldsymbol{R}_\mathrm{c2b}\boldsymbol{t} \\
    \boldsymbol{0} & 1
    \end{bmatrix}\\
    &=
    \begin{bmatrix}
    \boldsymbol{I} & \boldsymbol{R}_\mathrm{c2b}\boldsymbol{t} \\
    \boldsymbol{0} & 1
    \end{bmatrix}
\end{split}
\end{equation}
which is also translation-only. Thus, to obtain $a_t'$, we only need to add a translation $\boldsymbol{R}_\mathrm{c2b}\boldsymbol{t}$ to the translation component of $a_t$.

\subsection{Feature Matching with DINOv3}\label{subsec-apdx:feature-matching}

We derive feature-matching correspondence with DINOv3 by aggregating pixel features into pointcloud features in the following way: first, $N_\mathrm{img}=16$ images are rendered from different viewpoints of a shape (with colored texture) from the Library; then, $N_\mathrm{pcd}^0 = 5000$ points are sampled from the shape surface, which are subsequently filtered to remove points invisible from every rendered image (mostly points on the inner surfaces), resulting in about $N_\mathrm{pcd} = 3000$ valid points. Each valid point gets its final feature by averaging all pixel features retrieved by visible 2D projections of the point. For each designated point on a source shape $S_i$, we use the feature of its spatially closest neighbor from the sample pointcloud of $S_i$ to query that of $S_j$ and accept directly the closest neighbor in feature space as the matched point.

\section{Real-world Experiments}

\subsection{Task Definition}\label{subsec-apdx:task-definition}

We design 4 tasks for evaluating the effectiveness of ShapeGen. The tasks are described as follows:
\begin{itemize}
    \item \verb|hang_mug|. The gripper grasps the mug by its rim, transports it and plugs the mug handle into the rack.
    \item \verb|hang_mug_hard|. The gripper grasps the mug by its rim, \textbf{rotates} and transports it, then plugs the mug handle into the rack.
    \item \verb|serve_kettle|. The gripper grasps the kettle by its handle, rotates and transports it, then puts it onto a wooden block.
    \item \verb|pour_water|. The gripper grasps the kettle by its handle, transports it then tilts it for pouring water into a static mug.
\end{itemize}

\subsection{Evaluation Protocol}\label{subsec-apdx:evaluation-protocol}

In evaluation, we identify an execution a success if every step of the task is completed. If the policy fails to grasp the relevant object, or drops it amidst execution, then it is identified as a failure. Beside these, there are also task-specific requirements:
\begin{itemize}
    \item \verb|hang_mug| and \verb|hang_mug_hard|. The mug handle should be inserted onto the rack.
    \item \verb|serve_kettle|. The kettle should not drop down when the gripper releases.
    \item \verb|pour_water|. The vertical projection of the spout orifice should be within the inner diameter of the mug's rim.
\end{itemize}
Cases not fulfilling these requirements will also be identified as failure.

\begin{figure*}
    \centering
    \includegraphics[width=\linewidth]{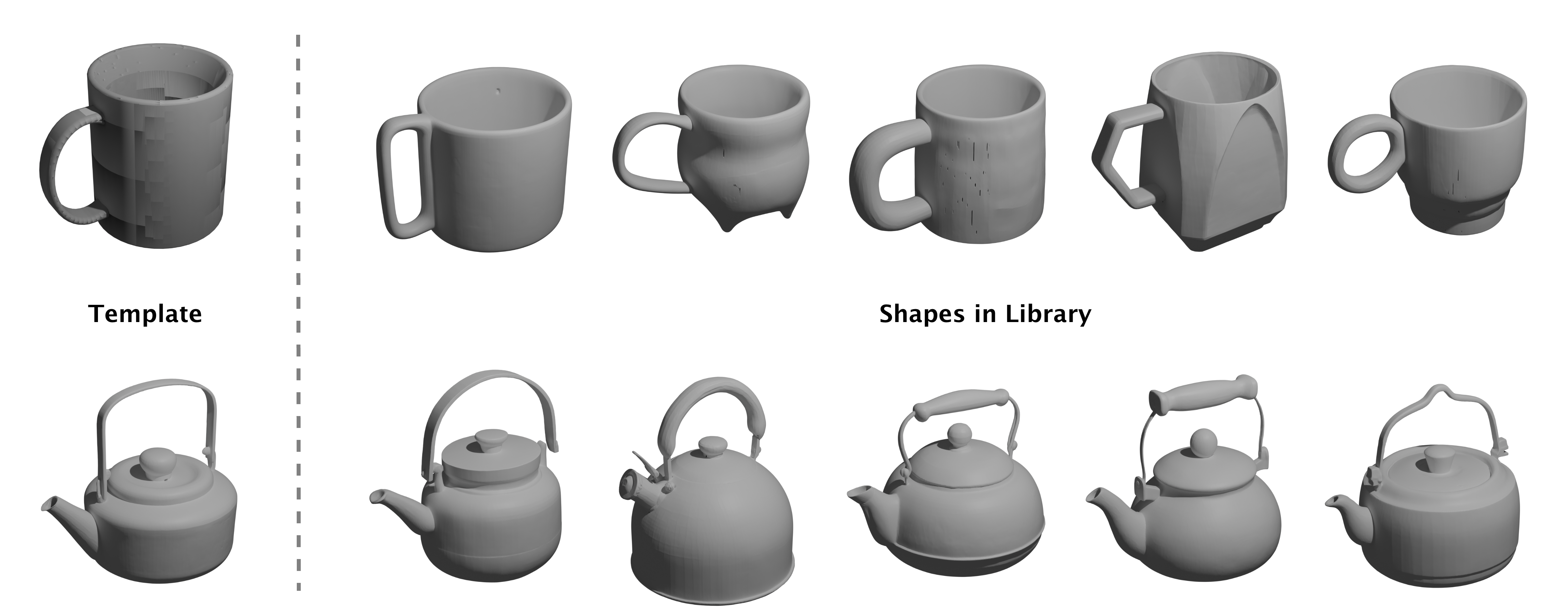}
    \caption{\textbf{Shapes in Library used for data generation.} The leftmost column shows the shape reserved as common template, while other 5 columns show shapes used to generate novel data.}
    \label{fig:shapes-in-library}
\end{figure*}

\begin{figure*}
    \centering
    \includegraphics[width=\linewidth]{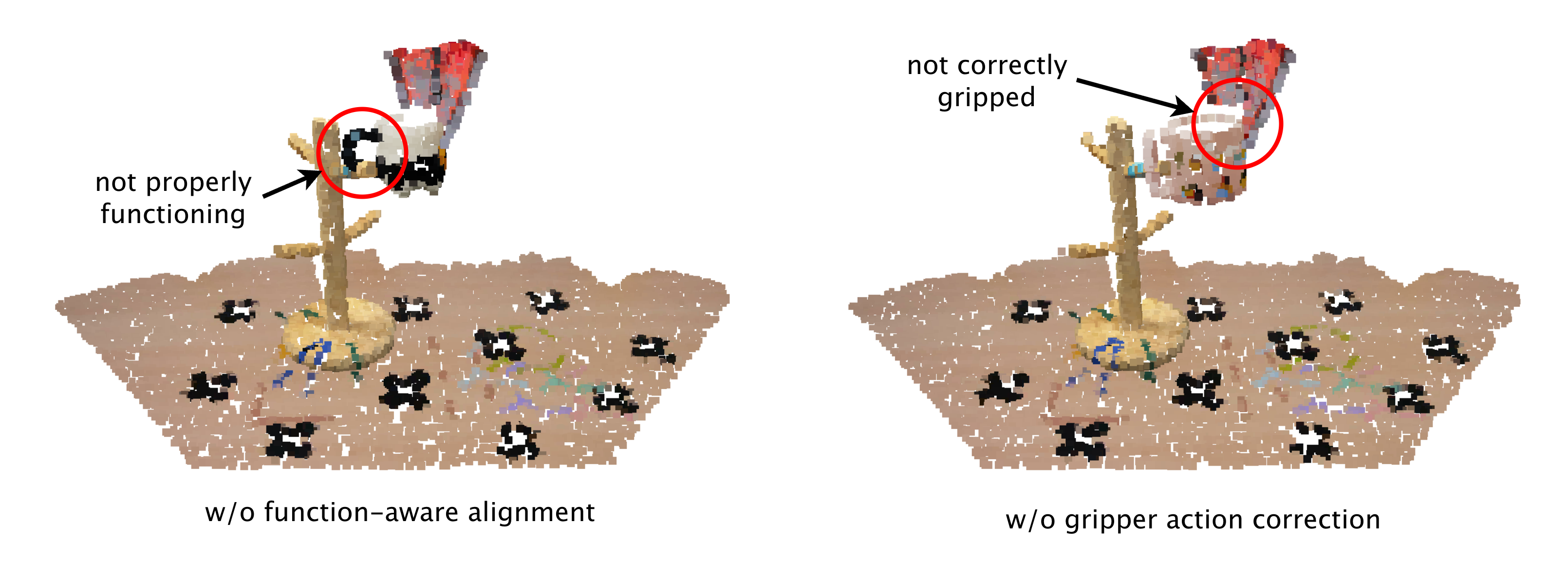}
    \caption{\textbf{Visualization of generation results with system component omission.} Omitting f.w.a. leads to misalignment, while omitting g.a.c. causes wrong gripping action in the case shown.}
    \label{fig:ablation-vis}
\end{figure*}

\subsection{Data Generation}\label{subsec-apdx:data-generation-detail}

For each task, we generate 15x more data with ShapeGen, each with a different shape. These 15x shapes are actually derived from 5 different 3D models, each at 3 distinct scales. The models are generated via SAM3D \cite{chen2025sam3d} from Internet-sourced images. Scales are selected based on the actual scale of source object. Let $l_\mathrm{src}$ be diagonal length of the axis-aligned bounding box of source object shape $S_\mathrm{a}$, we scale shapes in the Library such that their bounding box diagonal length equals to $0.8l_\mathrm{src}, l_\mathrm{src}$ and $1.2l_\mathrm{src}$ respectively. We visualize in detail the selected shapes as well as the picked template shape in Fig. \ref{fig:shapes-in-library}.

\subsection{Analysis of SR Variance}\label{subsec-apdx:SR-variance}

\begin{table}[htbp]
    \centering
    \caption{\textbf{Analysis of SR Variance.} Success rate is reported for the task \texttt{hang\_mug}.}
    \resizebox{0.98\linewidth}{!}{
        \begin{tabular}{c|ccccc|ccccc}
        \toprule
        \textbf{instance} & \multicolumn{5}{c|}{mug\_blackandwhite (\#1)} & \multicolumn{5}{c}{mug\_white (\#2)} \\
        \midrule
        \textbf{position} & \#1   & \#2   & \#3   & \#4   & \#5   & \#1   & \#2   & \#3   & \#4   & \#5 \\
        \midrule
        \textbf{SR/real} & 0/5   & 0/5   & 0/5   & 0/5   & 0/5   & 0/5   & 0/5   & 0/5   & 0/5   & 0/5 \\
        \textbf{SR/real+ShapeGen} & \textbf{5/5} & \textbf{5/5} & 0/5   & 0/5   & \textbf{5/5} & 0/5   & \textbf{5/5} & \textbf{5/5} & 0/5   & \textbf{5/5} \\
        \bottomrule
        \end{tabular}%
    }
    \label{tab:analysis-SR-variance}
\end{table}

We cut down on the number of trials for an observation that same initial configuration leads to highly repeatable outcomes. This can be demonstrated by results shown in Table \ref{tab:analysis-SR-variance}, obtained by running multiple rollouts on the same initial configuration (with full reset to inject inevitable small variations) of the task \verb|hang_mug|. The method by which most existing manipulation research extends evaluation rollouts, i.e. randomizing spatial configuration, does not suit our work because our primary concern is augmenting shape generalizability.

\section{Ablation on System Components}\label{sec-apdx:ablation}

\begin{table}[htbp]
  \centering
  \caption{\textbf{Ablation on system components.} Success rate is reported.}
  \resizebox{0.98\linewidth}{!}{
    \begin{tabular}{l|ccccc}
    \toprule
    \multicolumn{1}{c|}{\multirow{2}[4]{*}{\textbf{method}}} & \multicolumn{5}{c}{\textbf{hang\_mug}} \\
\cmidrule{2-6}          & \#1   & \#2   & \#3   & \#4   & \cellcolor[rgb]{ .804,  .953,  .976}total \\
    \midrule
    \textbf{5 source} & 0/5   & 0/5   & 0/5   & 1/5   & 1/20 \\
    \textbf{+ShapeGen w/o f.w.a.} & 0/5   & 0/5   & 0/5   & 0/5   & 0/20 \\
    \textbf{+ShapeGen w/o g.a.c.} & 1/5   & 0/5   & 0/5   & \textbf{3/5} & 4/20 \\
    \textbf{+ShapeGen} & \textbf{3/5} & \textbf{3/5} & \textbf{1/5} & 2/5   & \textbf{9/20} \\
    \bottomrule
    \end{tabular}%
    }
  \label{tab:ablation-system-component}%
\end{table}%

We conduct ablation study on ShapeGen system components with the task \verb|hang_mug|. Specifically, we study the influence of function-aware alignment and gripper action correction on performance of policies trained with ShapeGen-augmented data. In the \textbf{ShapeGen w/o f.w.a.} modification, we remove the function-aware alignment process, solely aligning the shapes to a common canonical orientation (which is equivalent to setting the warping to a uniform scaling), and carry out other procedures, including gripper action correction, as usual. In the \textbf{ShapeGen w/o g.a.c.} modification, we omit gripper action correction and keep all other procedures unchanged. Experimental result presented in Table \ref{tab:ablation-system-component} shows that removing either component results in significant performance drop. As shown in Fig. \ref{fig:ablation-vis}, omitting f.w.a. means manipulated objects in generated demos are not properly aligned for fulfilling their functionalities, which in our experiments results in that the mug being manipulated frequently collides with the rack, and that at times the gripper fails to pick up the mug altogether. Removing g.a.c. also leads to the latter problem, and at the same time means that all generated demos share the same action trajectory with the source demo. We observe in experiment that policies trained on ShapeGen w/o g.a.c. data fail in patterns similar to those trained with source data only.

\end{document}